%% file: root.tex
\documentclass[letterpaper, 10 pt, conference]{ieeeconf}

\IEEEoverridecommandlockouts
\usepackage{cite}
\usepackage{amsmath,amssymb,amsfonts}
\usepackage{algorithmic}
\usepackage{graphicx}
\usepackage{textcomp}
\usepackage{xcolor}
\usepackage{svg}
\usepackage{lipsum}  
\usepackage[thinc]{esdiff}
\usepackage{mathtools}
\usepackage{cleveref}
\usepackage{textcomp}
\usepackage{tabularx}  
\usepackage{siunitx}
\usepackage{caption}
\usepackage{subcaption}
\usepackage{upgreek}
\DeclareMathOperator*{\argmin}{arg\,min}
\DeclareMathOperator{\arctantwo}{arctan2}
\renewcommand*\descriptionlabel[1]{\hspace\leftmargin$#1$}

\title{\LARGE \bf Flying Trapeze Act Motion Planning Algorithm for Two-Link Free-Flying Acrobatic Robot}

\author{Thanapong Chuangyanyong$^{1}$, Panusorn Chinsakuljaroen$^{1}$,\\ Worachit Ketrungsri$^{1}$, and Thanacha Choopojcharoen$^{1,2}$
\thanks{$^{1}$All authors are associated with the Institute of Field Robotics, King Mongkut's University of Technology Thonburi, 126 Pracha Uthit Rd., Bang Mot, Thung Khru, Bangkok 10140.}%
\thanks{$^{2}$Thanacha Choopojcharoen is currently an adjunct lecturer at the Institute of Field Robotics, King Mongkut's University of Technology Thonburi.}%
}

\begin{document}

\maketitle
\thispagestyle{empty}
\pagestyle{empty}

\begin{abstract}
A flying trapeze act can be a challenging task for a robotics system since some act requires the performer to catch another trapeze or catcher at the end after being airborne. The objective of this paper is to design and validate a motion planning algorithm for a two-link free-flying acrobatic robot that can accurately land on another trapeze after free-flying in the air. First, the proposed algorithm plan the robot trajectory with the non-linear constrained optimization method. Then, a feedback controller is implemented to stabilize the posture. However, since the spatial position of the center-of-mass of the robot cannot be controlled, this paper proposes a trajectory correction scheme that manipulates the robot's posture such that the robot is still able to land on the target. Lastly, the whole algorithm is validated in the simulation that mimics real-world circumstances.
\end{abstract}

\input{src/Introduction.tex}

\input{src/ProblemFormulation}

\input{src/Modelling.tex}

\input{src/ControlStrategy.tex}

\input{src/Simulation.tex}

\section{Discussion and Conclusion}
The motion planning algorithm for a flying trapeze act has been designed for the two-link free-flying robot in this paper. The hybrid dynamics of the robot are derived and used in motion planning, controller synthesis, and simulation. The uncontrollability of the center-of-mass of the multibody system has been overcome by correcting the old trajectory in mid-air. By doing that, the ``success region" of the releasing angular position and the releasing angular velocity has increased substantially. The simulation results also reveal that the controller with all functionalities enabled yield $\SI{91.33}{\percent}$ success rate based on 300 simulations with uncertain parameters.

This paper also shows how the initial releasing condition affects the act. As shown in Fig. \ref{fig:region_plot}, the region with the disabled trajectory correction ($\text{TC}=0$) resulted in a smaller area than the enabled one ($\text{TC}=1$). However, when looking at the nominal releasing condition, it is vastly biased compared to the whole region. Therefore, if the performance of the gripper is modeled, the releasing condition can be placed accordingly to increase the possibility of performing a successful attempt. 

The control algorithm presented in this paper used the trajectory correction, which requires a more powerful processor to solve non-linear constrained optimization problems, which may be potentially more costly and more challenging to design and protect the physical hardware. Another potential setback is that the initial planning may take considerable time to find a feasible trajectory. 
Since the trapeze act is often a prepared routine act, the result from the previous computation can be reused, which the computation time can be neglected.
However, unlike the previous research, the proposed method incorporates various adjustable physical parameters such as target's position, joint limits, velocity limit, actuator limit, and angle of attack. In addition, the formulation of the proposed algorithm also allows the possibility of extension to a higher degree of freedom and flexibility in robot configuration, such as the gripper position and opening.

\bibliographystyle{IEEEtran}
\bibliography{root.bib}

\end{document}

%% file: src/Introduction.tex
\section{Introduction}
In recent years, robotics has become an integral part of immersive entertainment. New technologies allow robots to achieve complex behavior, one being a stunt double for untethered aerial performance in an amusement park. The most prominent example is Disney's ``Stuntronics," a performance robot that is being used at Walt Disney World\textsuperscript{\textregistered}, which is an implementation of the research from Disney Research, \cite{pope-2018}, where a three-link robot is designed to perform a safe landing from a swinging pendulum. Their work presents the potential of using robots to perform other complex acrobatics treats. This research will be focusing on designing a motion planning and control algorithm for one of the most prevalent aerial acrobatic performances, the flying trapeze act. The act involves a performer jumping from a platform while holding onto a flying trapeze, then catching another performer, also known as a catcher. Sometimes an experienced performer releases the trapeze and becomes airborne before catching the target. Nakanishi and Fukuda \cite{nakanishi-2000} proposed a control strategy for a ricochetal brachiation robot, which resembles a flying trapeze act. The controller works by combined system dynamics and mechanical energy regulator. However, their approach only works on targets that are the same height as the releasing height, which is rarely the case in the flying trapeze act. However, Wan et al. \cite{wan-2015} solve the aforementioned shortcoming by using a control method that combined the virtual constraint-based trajectory planning and tracking control algorithm. Both \cite{nakanishi-2000} and \cite{wan-2015} are able to achieve accuracy of the final catching position. Especially \cite{nakanishi-2000}, which can adjust the robot posture when the releasing condition (the spatial position and velocity when the robot starts to experience free fall) is off from the desired value. The posture adjustment is essential since the system's center-of-mass follows a parabolic trajectory predetermined by the releasing condition. For the research to move closer to a physical realization, the control algorithm should be taken into account the adjustable target's position similar to \cite{wan-2015}, and it should incorporate ``posture adjustment" similar to \cite{nakanishi-2000} in case of control delay or physical uncertainties. Additionally, there are several points that must be addressed, for instance, the robot's physical limits (joint limits, actuator torque, and actuator velocity) and the attack angle of which the robot grasp the target. 

The algorithm will be implemented on an acrobatic robot research platform named Mon$\chi$. Although Mon$\chi$ is designed for general acrobatic movement, this paper only focuses on the flying trapeze act. Mon$\chi$ consists of two rigid links connected via an actuated revolute joint. At the end of one of the links lies a gripper module used to clutch the trapeze and land on the target. The robot is specifically designed to stay on the sagittal plane closely. Therefore, while the robot is flying in the air, there are four degree-of-freedom (DOF) of movement with only a single-DOF actuator. Moreover, when the robot clutches the trapeze, there are two-degree-of-freedom with the same actuator. This kind of mechanism is called an underactuated system, which adds more challenges to the design of the algorithm.

%% file: src/ProblemFormulation.tex
\section{Problem Formulation}
A flying trapeze act requires a positional accuracy at the end of the trajectory to be able to catch the next trapeze or the catcher. The tolerance of the distance between the final landing position and the target is aimed to be less than $\SI{5}{\centi\meter}$ according to the ratio between the average human hand and the average length of the human body with extended arms. The robot, Mon$\chi$, consists of two rigid links. The first link provides a mounting point for an embedded computer, a power source, an actuator, an inertial measurement unit (IMU), and a gripper module. The second link is connected to the first link via an actuated revolute joint. $F_g$ is the coordinate frame of the landing position attached to the first link. The origin of $F_g$ is the contact point that the motion planner will attempt to navigate to the target bar. The z-axis of $F_g$ is parallel to the axis of actuation, and the x-axis is perpendicular to the contact surface of the gripper.

To verify the motion planning algorithm of the flying trapeze act, the apparatus must be designed accordingly. The apparatus comprises a starting platform, a trapeze, and a target bar. The target bar acts as a catcher; however, it is designed to be stationary rather than swingable. The function of the starting platform is to lock Mon$\chi$ in position until the starting signal is set, which then instantly releases the robot. A global coordinate frame $F_0$ has its z-axis coinciding with the axis of rotation of the trapeze’s hinge, and its y-axis points upward—the gravity points in the $-y$ direction of the $F_0$. The x-axis is parallel to the ground and points to the target bar. The target bar is adjustable in the sagittal plane, and $ F_t$ denotes its attached coordinate frame. 

\begin{figure}[htbp]
    \centering
    \includegraphics[width=\textwidth/2-0.5cm]{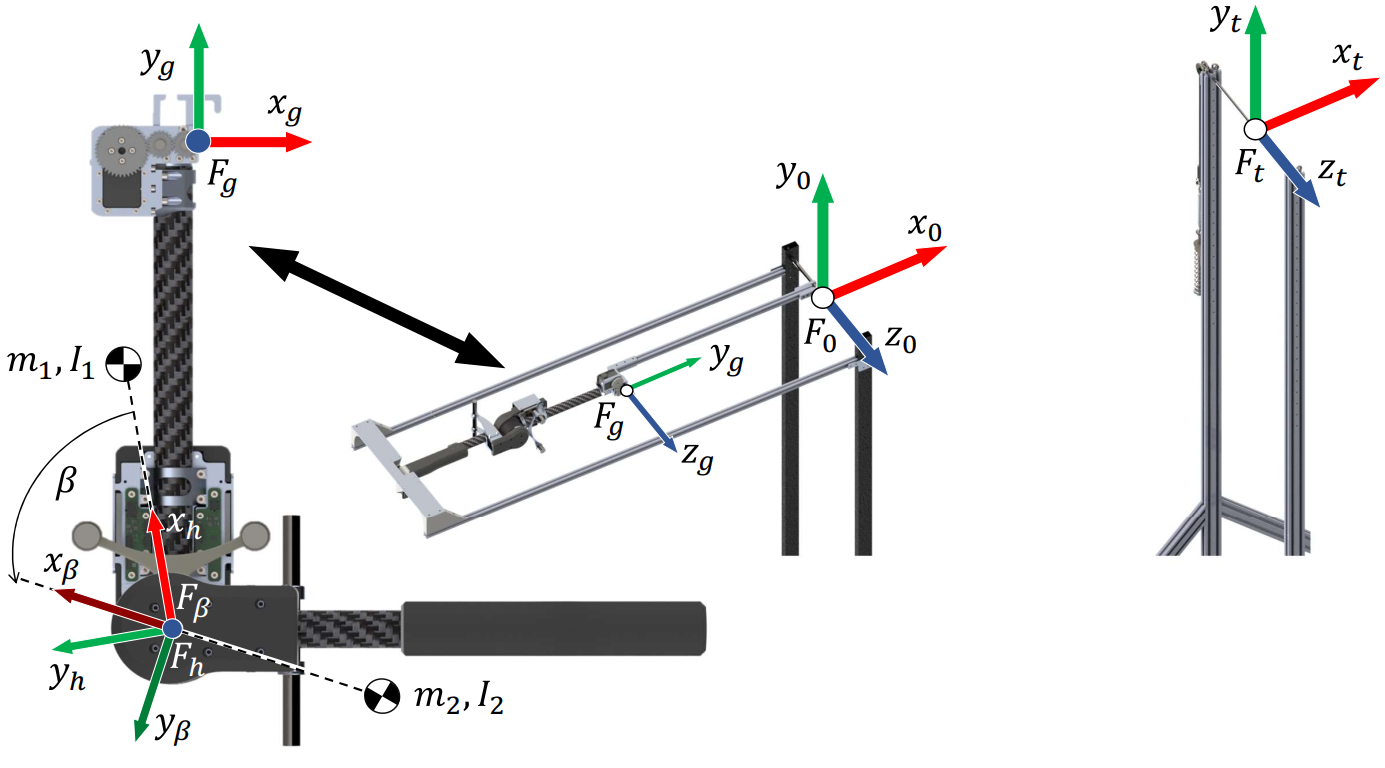}
    \caption{Coordinate frames of Mon$\chi$ and its environment.}
    \label{fig:monx_model_super}
\end{figure}

%% file: src/Modelling.tex
\section{Modeling}
A trapeze act consists of two phases: a ``swinging phase" and a ``flying phase." The robot is in a swinging phase when it clutches the trapeze with its gripper module. Once the robot releases its gripper from the trapeze, it transitions into a flying phase. By denoting the releasing time by $t_{rel}$, the transition occurs only once when time $t$ is equal to $t_{rel}$. Since there are only two phases with a simple transition behavior, equations of motion are derived separately. The system's mode can be switched with a discrete state variable $s\in\{0,1\}$. When $s$ is $0$, the robot is in the swinging phase. When $s$ is $1$, the robot is in the flying phase. 

\begin{figure}[htbp]
    \centering
    \includegraphics[width=\textwidth-12cm]{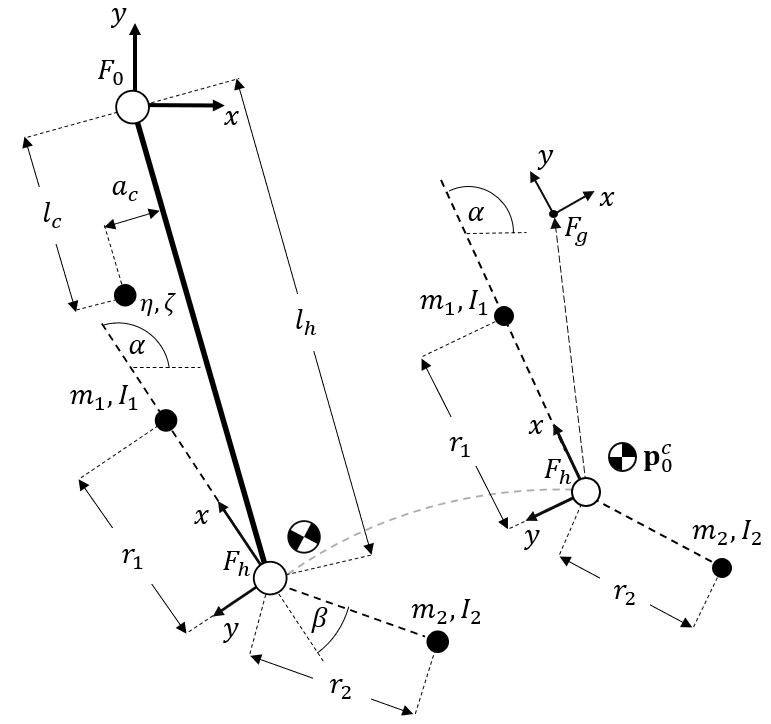}
    \caption{Mon$\chi$'s model in the swinging phase (left) and Mon$\chi$'s model in the flying phase (right).}
    \label{fig:monx_model}
\end{figure}

\subsection{Flying Phase}
Fig.~\ref{fig:monx_model} shows the model of the robot. $m_1, m_2, I_1, I_2$ are the mass and moment of inertia about the local z-axis of the first link and the second link, respectively. $F_h$ is the hip's coordinate frame,  which is rigidly attached to the first link. Its origin coincides with the axis of actuation, and the x-axis points directly toward the center-of-mass of the first link. $r_1$ and $r_2$ are the distance corresponding to the first link center-of-mass measured from the origin of $F_h$ in the $+x$ direction and the distance corresponding to the second link center-of-mass measured from the origin of $F_\beta$ in the $-x$ direction, respectively.

In the flying phase, given negligible air drag, Mon$\chi$ has no external wrenches acted on it besides the gravity. Consequently, the center-of-mass trajectory is modeled as a parabolic trajectory \eqref{eq:com_traj} and will be entirely determined by the position and the velocity right after it leaves the trapeze. Thus, the angular position and angular velocity of both links can be decoupled from its center-of-mass position and velocity \cite{dubowsky-1993}. The spatial position and velocity can be added in the form of a simple time-varying variant of projectile motion afterward. Therefore, the generalized coordinate ($\mathbf{q}$) consists of two configuration variables: a rotation of the first link relative to the x-axis of $F_0$ about the z-axis of $F_0$ ($\alpha$), and a rotation of the second link relative to the x-axis of $F_h$ about the z-axis of $F_h$ ($\beta$). The Euler-Lagrange equation is used to obtain the equation of motion.

\subsection{Swinging Phase}
In the swinging phase, Mon$\chi$ is modeled as a double pendulum with an actuated revolute joint on the second joint. The length of the first link of the double pendulum ($l_h$) is measured from the origin of $F_0$ to the origin of $F_h$ while the robot clutches the trapeze. $\eta$ is the mass of the first link with the trapeze. $\zeta$ is the moment of inertia about the z-axis of the first link while both links are rigidly coupled. One end of the trapeze is attached to the origin of $F_0$ with a revolute joint, as seen in Fig. \ref{fig:monx_model}. $\kappa_1$ and $\kappa_2$ are the angular offset measured from the x-axis of $F_h$ to the rotating axle of the trapeze, and the angular offset measured from -x-axis of $F_\beta$ to the center-line of the second physical link, respectively. $\sigma_\alpha$ and $\sigma_\beta$ are the damping coefficient of the revolute joint at $F_0$ and $F_h$, respectively. The former is added to maintain continuity while transitioning between phases. The latter is added to determine the physical limit of the actuated joint. Similar to the flying phase, equation motion is formulated; however, the structure is identical. As a result, the equation of motion can be expressed such that the only difference is a parameter vector ($\pmb{\uptheta}_s$). Parameters are separated from both equations of motion, which can now be written as in \eqref{eq:man_eq}.

\begin{equation}\label{eq:man_eq}
    \mathbf{\uptau} = \mathbf{M}_s(\mathbf{q})\ddot{\mathbf{q}}+\mathbf{C}_s(\dot{\mathbf{q}}, \mathbf{q})\dot{\mathbf{q}}+\mathbf{B}_s\dot{\mathbf{q}}+\mathbf{g}_s(\mathbf{q})
\end{equation}

\begin{align}\label{eq:M_s}
    \mathbf{M}_s(\mathbf{q}) =
    \begin{bmatrix}
        2\theta_{s,1}\cos(\beta)+\theta_{s,2}+\theta_{s,3}& \theta_{s,1}\cos(\beta)+\theta_{s,3}\\
        \theta_{s,1}\cos(\beta)+\theta_{s,3} & 
        \theta_{s,3}
    \end{bmatrix}
\end{align}

\begin{equation}\label{eq:C_s}
    \mathbf{C}_s(\dot{\mathbf{q}}, \mathbf{q}) = 
    \begin{bmatrix}
        -\theta_{s,1}\sin(\beta)\dot{\beta} & -\theta_{s,1}\sin(\beta)(\dot{\alpha}+\dot{\beta}) \\
        \theta_{s,1}\sin(\beta)\dot{\alpha} & 0
    \end{bmatrix}
\end{equation}

\begin{align}\label{eq:G}
    \mathbf{g}_s(\mathbf{q}) = 
    \begin{bmatrix}
        \theta_{s,6}\sin(\alpha')+\theta_{s,5}\cos(\alpha')+\theta_{s,4}\cos(\alpha'+\beta)\\
        \theta_{s,4}\cos(\alpha'+\beta)
    \end{bmatrix}
\end{align}

\begin{equation*}
    \mathbf{B}_s = 
    \begin{bmatrix}
        \theta_{s,7} & 0 \\
        0 & \theta_{s,8} \\
    \end{bmatrix},~\alpha' = \alpha + \kappa_1
\end{equation*}

\begin{equation}\label{eq:theta_0}
    \pmb{\uptheta}_0 =
    \begin{bmatrix}
        m_2{l_h}{r_2} \\
        \eta(l_{c}^2+a_{c}^2)+(\eta+m_2)l_{h}^2-2\eta l_{h} l_{c}+\zeta \\
        m_2{r_2^2}+I_2 \\
        \eta{g_y}a_{c} \\
        \eta{g_y}(l_h-l_{c})+m_2{g_y}l_h \\
        m_2{g_y}r_2 \\
        \sigma_\alpha \\
        \sigma_\beta \\
    \end{bmatrix}
\end{equation}

\begin{equation}\label{eq:theta_1}
    \pmb{\uptheta}_1 =
    \begin{bmatrix}
        \mu r_1 r_2 \\
        \mu r_1^2+I_1 \\
        \mu r_2^2+I_2 \\
        0 \\
        0 \\
        0 \\
        0 \\
        \sigma_\beta \\
    \end{bmatrix},~~\mu=\frac{m_1 m_2}{m_1+m_2}
\end{equation}

\subsection{State Space Representation}
Finally, the state space representation of the system can be written as presented in \eqref{eq:ss_var} and \eqref{eq:ss_var_dot}.

\begin{equation}\label{eq:ss_var}
    \mathbf{x} =
    \begin{bmatrix}
        \mathbf{q} \\
        \dot{\mathbf{q}} \\
    \end{bmatrix}
\end{equation}

\begin{equation}\label{eq:ss_var_dot}
    \dot{\mathbf{x}}=f_{\lambda(t)}(\mathbf{x},u),~
    \lambda(t) = 
    \begin{cases}
        0, & \text{if}~t \leq t_{rel}\\
        1, & \text{otherwise.}\\
    \end{cases}
\end{equation}

Where:

\begin{gather}\label{eq:ss_eq}
    f_s(\mathbf{x}, u) = \nonumber\\
    \begin{bmatrix}
        \dot{\mathbf{q}}\\
        \mathbf{M}_s(\mathbf{q})^{-1}\left(
        \begin{bmatrix}
            0\\
            1\\
        \end{bmatrix}u
        -\mathbf{C}_s(\dot{\mathbf{q}}, \mathbf{q})\dot{\mathbf{q}}-\mathbf{B}_s\dot{\mathbf{q}}-\mathbf{g}_s(\mathbf{q})
        \right)\\
    \end{bmatrix}
\end{gather}

%% file: src/ControlStrategy.tex
\section{Control Strategy}
One of the most prevalent motion planning methods for an underactuated system like Mon$\chi$ is non-linear constrained trajectory optimization. The optimization approach can be constrained to satisfy the constraints stated above. Moreover, the system only has four state variables, which further reduces optimization time. Thus, the strategy is to find an optimal trajectory for Mon$\chi$ to follow. Furthermore, the uncertainty in the robot's physical properties can cause the robot to miss the target bar even though the planned trajectory is feasible. Therefore, the controller shall be implemented to follow its reference pose trajectory. 

\subsection{Optimal Motion Planning}
To land at the target bar, various constraints must be considered in this optimization problem. First, the landing point of the robot ($\mathbf{p}_0^g$) must be at the target bar ($\mathbf{p}_0^t$) at the end of the trajectory. The landing point can be transformed from the center-of-mass position ($\mathbf{p}_0^c$), which can be computed from \eqref{eq:com_traj}. Every position vector in this paper is constructed with respect to $F_0$.

\begin{equation}\label{eq:p_0c_to_p_0p}
    \mathbf{p}_0^g(\mathbf{q},t)=\mathbf{p}_0^c(t)+\mathbf{p}_c^h(\mathbf{q})+\mathbf{p}_h^g
\end{equation}

\begin{equation}\label{eq:p_ch}
    \begin{split}
        \mathbf{p}_c^h(\mathbf{q}) &=\frac{m_1r_1}{m_1+m_2}
        \begin{bmatrix}
        \cos(\alpha)\\
        \sin(\alpha)\\
    \end{bmatrix}\\ & +\frac{m_2r_2}{m_1+m_2}
    \begin{bmatrix}
        \cos(\alpha+\beta)\\
        \sin(\alpha+\beta)\\
        \end{bmatrix}
    \end{split}
\end{equation}

$\mathbf{p}_h^g$ is the position of the landing point frame ($F_g$) relative to the origin of $F_h$. $\mathbf{p}_h^g$ is used here rather than translating directly from the center-of-mass to the landing position because, this way, the landing position is always referenced to the robot's hip, which is easier for development purposes. 

Next, the landing point must not be obstructed by any part of the robot. For this research, the landing point has an opening angle between $\gamma_{min}$ and $\gamma_{max}$ with respect to the x-axis of the $F_g$. If the angle of the position vector $\mathbf{p}_g^t$ while the robot is approaching lies within the opening with respect to the x-axis of $F_g$ \eqref{eq:psi_func}, the landing point will not be obstructed. Moreover, an offset of $2\pi\nu$ can be added to alter the final posture; consequently, force the robot to somersault where $\nu$ is a number of somersault attempts.

\begin{equation}\label{eq:psi_func}
    \psi(\mathbf{x}(t)) = \alpha(t)+\mathbf{R}_0^g-\arctantwo(p_{g}^{t,y}(\mathbf{q},t), p_{g}^{t,x}(\mathbf{q},t))
\end{equation}

\begin{equation}\label{eq:p_pg}
    \mathbf{p}_g^t(\mathbf{q},t) = \mathbf{p}_0^t-\mathbf{p}_0^g(\mathbf{q},t)
\end{equation}

Lastly, the minimum effort squared is chosen to be the cost function. The aforementioned function usually produces a smooth and life-like trajectory. With these constraints, the optimization problem can be formulated as in \eqref{eq:offline_opt}.

\begin{equation}\label{eq:offline_opt}
    \langle\mathbf{x}^*(t), u^*(t), T_0^*, T_1^*\rangle =
    \argmin_{\mathbf{x}(t),\\ u(t),\\ T_0,\\ T_1} \int_{\tau=0}^{T_0+T_1} u^2(\tau) \mathrm{d}\tau
\end{equation}

subject to:
\begin{description}
    \item $\mathbf{x}(0)=\mathbf{x}_0$
    \item $\dot{\mathbf{x}}(t) = f_{\lambda(t)}(\mathbf{x}(t),u(t))\forall t \in[0,T_0+T_1]$ 
    \item $\gamma_{min}\leq\psi(\mathbf{x}(\tau))-2\pi\nu\leq\gamma_{max}\\ \forall\tau\in[T_0+\phi_{min}T_1,T_0+\phi_{max}T_1]$
    \item $\mathbf{p}_0^g(T_0+T_1) = \mathbf{p}_0^t$\\
\end{description}

$\mathbf{x}_0$ is the desired initial state, and $T_s$ is the trajectory duration. The final state of the swinging phase, $\mathbf{x}(t_{rel})$, is used to calculate the initial condition of the center-of-mass trajectory in \eqref{eq:p_0c_0} and \eqref{eq:v_0c_0} while the robot is in the flying phase. $\phi_{min}$ and $\phi_{max}$ are approaching constants that range between 0 and 1. $\phi_{min}$ is always lesser than $\phi_{max}$. If $\phi_{min}$ and $\phi_{max}$ are set to $0.75$ and $0.95$, respectively, this means the third constraint in \eqref{eq:offline_opt}, approaching constraint, will only be enforced from \SI{75}{\percent} to \SI{95}{\percent} of the flying phase trajectory in the time domain. Lastly, the boundary on state trajectory and control input is also added to ensure that the robot operates within its limit. 

\begin{equation}\label{eq:com_traj}
    \mathbf{p}_0^c(t) = \mathbf{p}_0^c(t_{rel}) + \mathbf{v}_0^c(t_{rel})t +
    \begin{bmatrix}
        0\\
        g_y\\
    \end{bmatrix} \frac{t^2}{2}
\end{equation}

\begin{equation}\label{eq:p_0c_0}
    \mathbf{p}_0^c(t_{rel}) = \mathbf{p}_0^h(t_{rel}) - \mathbf{p}_c^h(t_{rel})
\end{equation}

\begin{equation}\label{eq:v_0c_0}
    \mathbf{v}_0^c(t_{rel})  = \dot{\mathbf{p}}_0^h(t_{rel})  - \dot{\mathbf{p}}_c^h(t_{rel}) 
\end{equation}

\begin{equation}\label{eq:v_0h_0}
    \mathbf{p}_0^h(t_{rel})  = -
\begin{bmatrix}
        l_{h}\cos(\alpha(t_{rel}) +\kappa_1)\\
        l_{h}\sin(\alpha(t_{rel}) +\kappa_1)\\
    \end{bmatrix}
\end{equation}

This motion planning algorithm is planned to be implemented on an embedded computer. Therefore, the optimization is parameterized beforehand. Some aspects of the problem must be modified. For instance, the Hermite-Simpson collocation method is used here to enforce the system dynamics \cite{hargraves-1987} rather than using a numerical integration method. The resulting state trajectory ($\mathbf{x}^*$) and resulting control input ($u^*$) will be stored as cubic coefficients and linear coefficients, respectively.

\begin{figure}[htbp]
    \centering
    \includegraphics[width={\textwidth/2-0.5cm}]{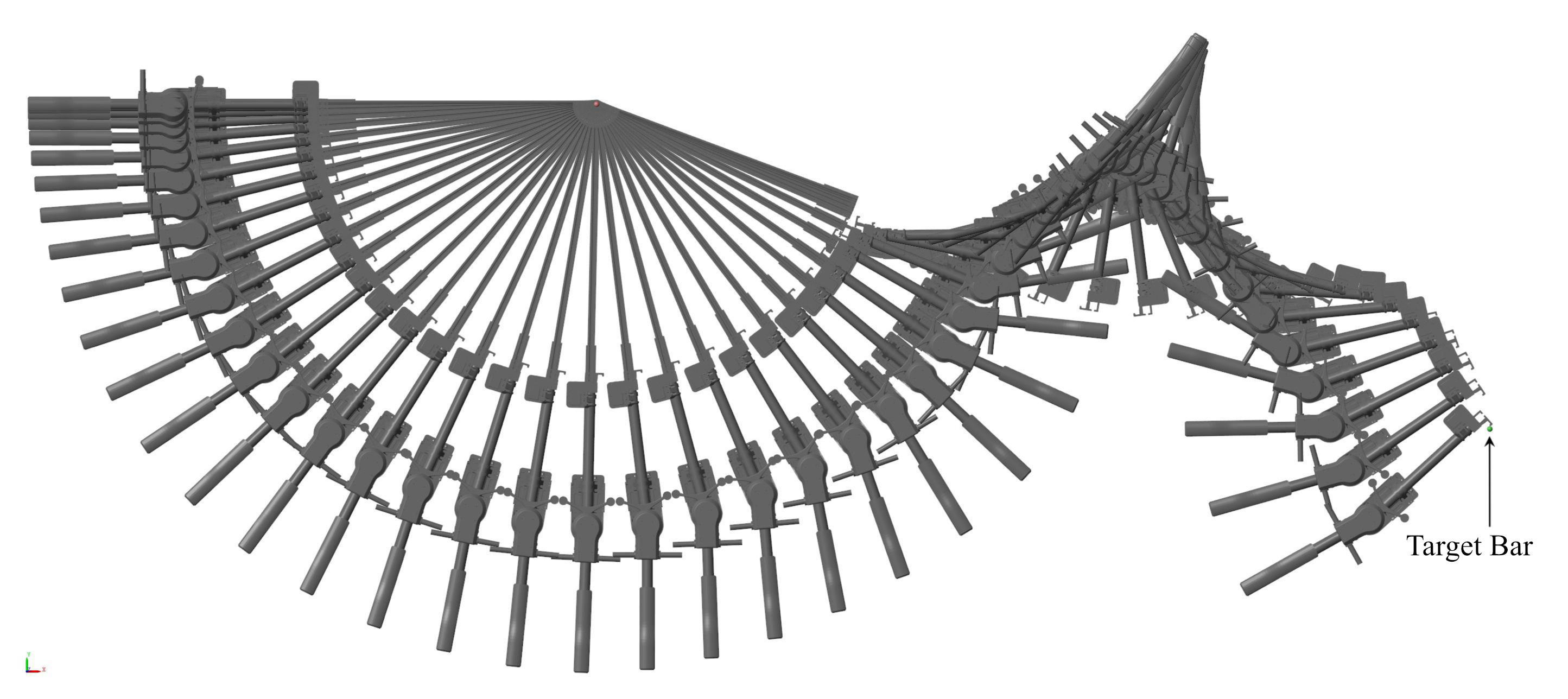}
    \caption{Example of a single somersault trajectory ($\nu=1$) from non-linear constrained optimization method, in this example, $\mathbf{p}_0^t=[2.44~-0.9]^\top$.}
    \label{fig:trajectory_result}
\end{figure}

\subsection{Posture Control}
After obtaining the feasible trajectory, Mon$\chi$ must stabilize its trajectory according to the reference trajectory. Note that the controller only has to stabilize the state variable ($\mathbf{x}$), not the position in the sagittal plane. A time-varying linear quadratic regulator (TVLQR) is used here since it has been shown to perform sufficiently in several physical nonlinear systems in terms of tracking performance, as seen in \cite{barry-2017}, \cite{moore-2016}, and \cite{farzan-2019}. Especially the last mentioned, which is highly similar to our system both in robot topology and the trajectory it is trying to follow.

\begin{equation}\label{eq:tvlqr}
    u(t) = u^*(t)-\mathbf{K}(t)(\mathbf{x}-\mathbf{x}^*(t))
\end{equation}

The controller stabilizes the linearized plant in its error coordinate \eqref{eq:tvlqr}. $\mathbf{K}(t)$ is a piece-wise linear optimal gain matrix. It can be solved by forming a differential-algebraic Riccati equation and integrating it backward in time. The final cost of the swinging phase can be set to the initial cost of the flying phase to make $\mathbf{K}(t)$ continuous. $\mathbf{K}(t)$ can also be stored in the same manner as the control input ($u^*$).

\subsection{Trajectory Correction}
The uncontrollability of Mon$\chi$'s center-of-mass can pose a potential problem. For instance, when the gripper misses the releasing time by little, the releasing position ($\mathbf{p}_0^c(t_{rel})$), and the releasing velocity ($\mathbf{v}_0^c(t_{rel})$) can vary from the reference values greatly. Consequently, the robot's center-of-mass trajectory will also change. This causes the gripper to miss the target because the posture stabilizer only stabilizes the state variable ($\mathbf{x}$). Even though the center-of-mass trajectory can not be altered without an external wrench, the gripper position ($\mathbf{p}_0^g(t)$); however, can be altered if the actual releasing state is known \cite{nakanishi-2000}. Though, the correction scheme in \cite{nakanishi-2000} only works for horizontal bars with the same height. Thus, the optimal control approach from earlier can be reduced to solve a new trajectory. The true releasing state can be estimated by using any online parameter estimator, e.g., a recursive least square. In this case, the least square estimator is used because the trajectory is relatively short. Then, the reduced version of the previous optimization problem can be used to find a new feasible trajectory. 

\begin{equation}\label{eq:online_opt}
    \langle\mathbf{x}^*(t), u^*(t), T^*\rangle =
    \argmin_{\mathbf{x}(t),\\ u(t),\\ T,\\ \upsilon} \int_{\tau = t_{rel}}^T u^2(\tau) \mathrm{d}\tau
\end{equation}

subject to:
\begin{description}
    \item $\mathbf{x}(0)=\hat{\mathbf{x}}(t_{rel})$
    \item $\dot{\mathbf{x}}(t) = f_{1}(\mathbf{x}(t),u(t))\forall t \in(t_{rel},T]$ 
    \item $\gamma_{min}\leq\psi(\mathbf{x}(\tau))+\upsilon\leq\gamma_{max}\forall\tau\in[\phi_{min}T,\phi_{max}T]$
    \item $\mathbf{p}_0^g(T) = \mathbf{p}_0^t$\\
\end{description}

Now, the problem is reduced to only optimizing the flying phase. $\upsilon$ is an auxiliary decision variable and is there, in place of $-2\pi\nu$, because the final posture is already constrained by the releasing state there. The time window between robot releasing and robot landing on the target is already determined by its center-of-mass trajectory; the robot also experienced the nonholonomic constraint on velocity emerges, in this case, from the conservation of angular momentum in a multibody system \cite{xin-2004}. Hence, the possible range of the final posture is already determined; adding more constraints on the final posture may cause the feasible trajectory to be more strenuous to find or even impossible to find. Most importantly, the flying portion of the previously optimized decision variables can be utilized as an initial guess to further decrease the optimization time. 

\begin{figure}[htbp]
    \centerline{\includegraphics[scale=0.6]{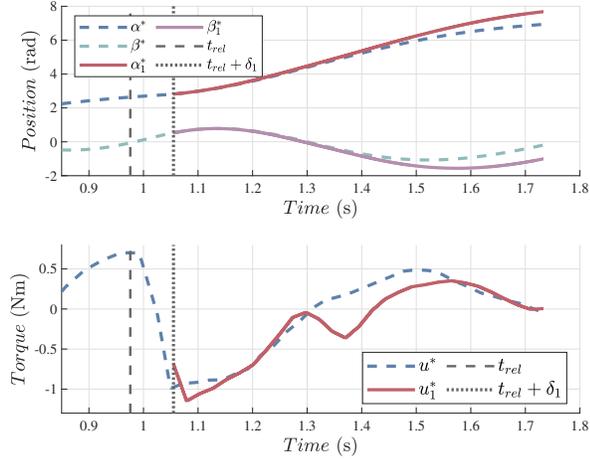}}
    \caption{Visualization of the new reference position trajectory ($\alpha_1^*$, $\beta_1^*$), and the new reference actuator's effort ($u_1^*$).}
    \label{fig:corrected_traj}
\end{figure}

\begin{figure}[htbp]
    \centerline{\includegraphics[width={\textwidth/2}-10mm]{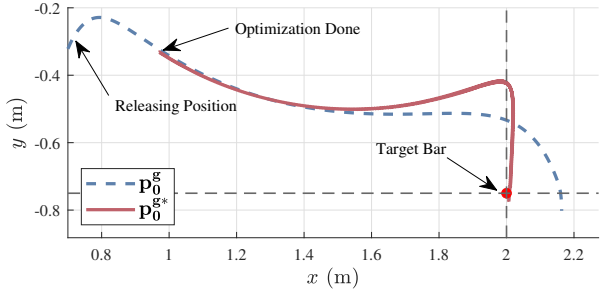}}
    \caption{Visualization of the corrected gripper trajectory ($\mathbf{p}_0^{g*}$)}
    \label{fig:corrected_p_0p}
\end{figure}

After the online optimization is finished, the posture trajectory is corrected by calculating the new optimal time-varying LQR gain. Fig. \ref{fig:corrected_traj} and Fig. \ref{fig:corrected_p_0p} show how trajectory correction affect the reference trajectory of the robot. The newly optimized state reference and control input reference can be denoted as $\mathbf{x}^*_i$, and $u^*_i$ instead, where $i$ is the number of the optimization attempt since the trajectory correction algorithm can be initiated as many times as specified. This may be helpful when the robot experiences another change in its center-of-mass trajectory, like a strong gust of wind. 

\begin{figure}[htbp]
    \centerline{\includegraphics[width = {\textwidth/2}]{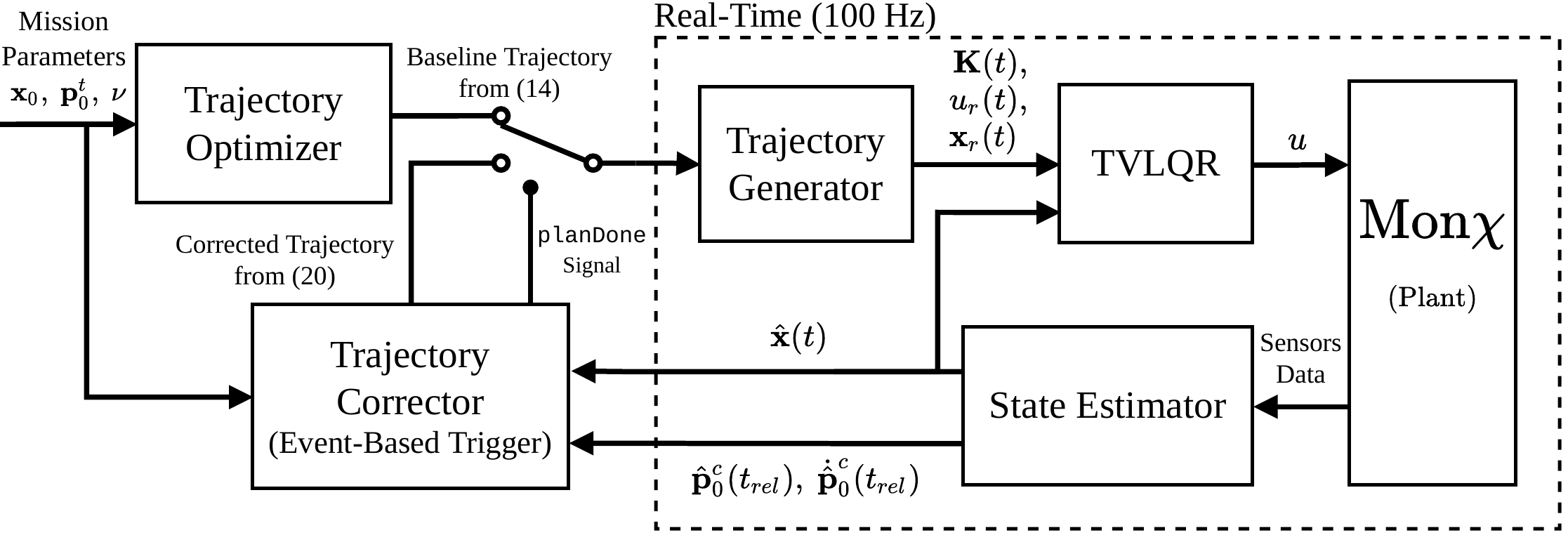}}
    \caption{A simplified block diagram of the proposed control scheme.}
    \label{fig:control_scheme}
\end{figure}

With all necessary components in place, the full control scheme can be realized. The simplified block diagram of the control scheme can be seen in Fig. \ref{fig:control_scheme}, which shows that only the state estimator, TVLQR, and the trajectory generator must be computed in a real-time manner. And other tasks that require solving nonlinear optimization problems should be isolated in another computing thread or processor.

%% file: src/Simulation.tex
\section{Simulation}
Both Mon$\chi$ and the testing environment are simulated with Simulink with a high level of detail, including the contact between every element.  

\subsection{Trajectory Correction Verification}
To verify the effectiveness of the trajectory correction algorithm, simulation is performed in MATLAB to numerically find the releasing conditions that result in a successful landing. Results are compared between two scenarios: trajectory correction algorithm disabled ($\text{TC}=0$) and trajectory correction algorithm enabled ($\text{TC}=1$). Note that all trajectory correction attempts must be done within the time-limit, in this experiment, \SI{0.5}{\second}. Below are parameters for this experiment and the next one, where $\mathbf{Q}$ and $\mathbf{R}$ are weighting matrices.

\begin{equation*}
    \mathbf{Q} = 
    \begin{bmatrix}
        4 & 0 & 0 & 0 \\ 
        0 & 2 & 0 & 0 \\
        0 & 0 & 2 & 0 \\
        0 & 0 & 0 & 2 \\
    \end{bmatrix},~
    \mathbf{R} = \begin{bmatrix}0.3\end{bmatrix}
\end{equation*}

\begin{equation*}
    \mathbf{p}_0^t = 
    \begin{bmatrix}
        1.9 \\ 
        -1.725 \\
    \end{bmatrix},~
    \mathbf{x}_0 =
    \begin{bmatrix}
        -\kappa_1 & \kappa_1 & 0 & 0\\
    \end{bmatrix}^\top
\end{equation*}

\begin{gather*}
    \gamma_{min} = \SI{-40}{\degree},~\gamma_{max}=\SI{10}{\degree}\\\phi_{min}=0.73,~\phi_{max}=0.90,~\nu=1
\end{gather*}

After the optimization process completed, a nominal trajectory and the releasing condition are obtained.

\begin{gather*}
        \alpha(t_{rel}) = \SI{2.834}{\radian},~\beta(t_{rel}) = \SI{0.390}{\radian}\\ 
        \dot{\alpha}(t_{rel}) = \SI{1.7816}{\radian\per\second},~\dot{\beta}(t_{rel}) = \SI{6.139}{\radian\per\second}\\ 
        t_{rel} = \SI{1.0204}{\second}
\end{gather*}

The releasing condition progresses incrementally away from the nominal releasing condition to construct a ``margin" of releasing conditions that result in successful acts.  

\begin{figure}[htbp]
    \centering
    \begin{subfigure}{.25\textwidth}
      \centering
      \includegraphics[scale = 0.38]{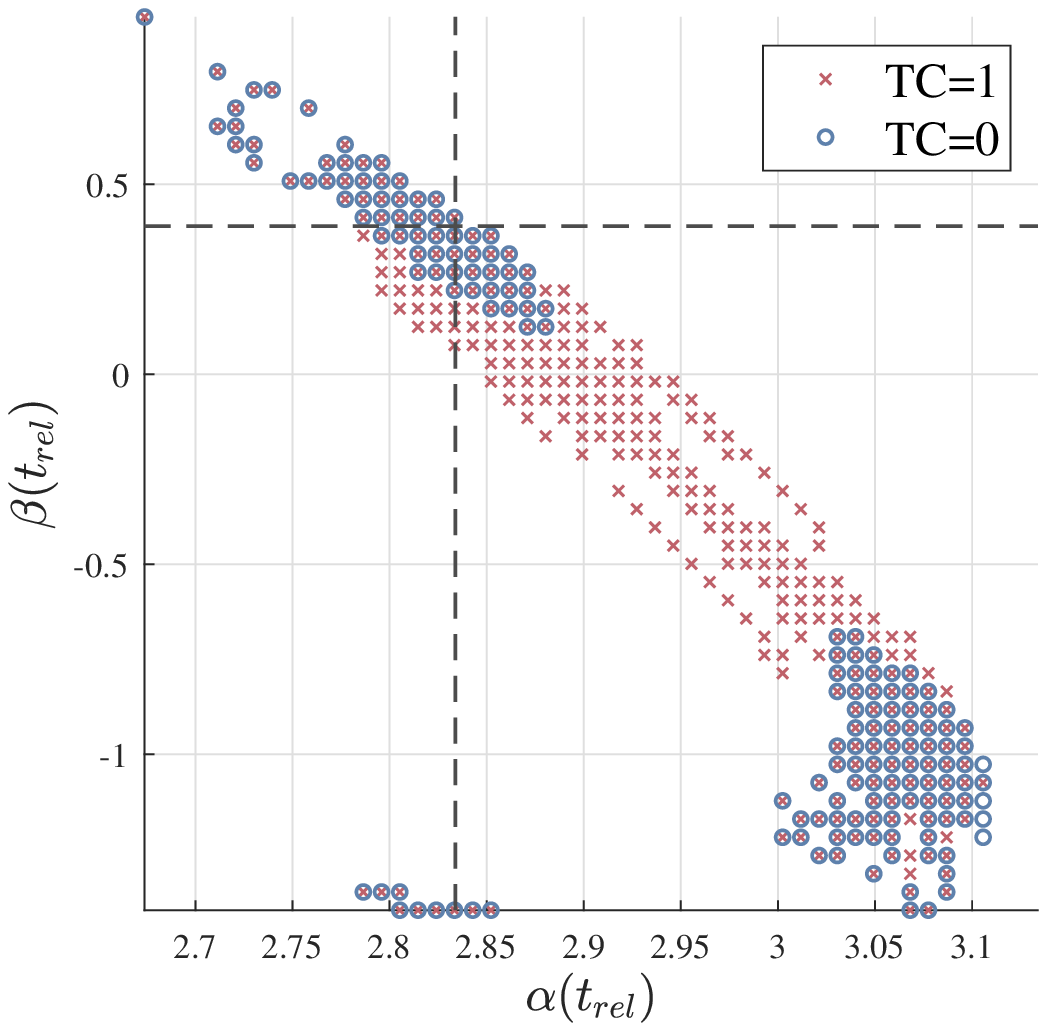}
    \end{subfigure}%
    \begin{subfigure}{.25\textwidth}
      \centering
      \includegraphics[scale = 0.38]{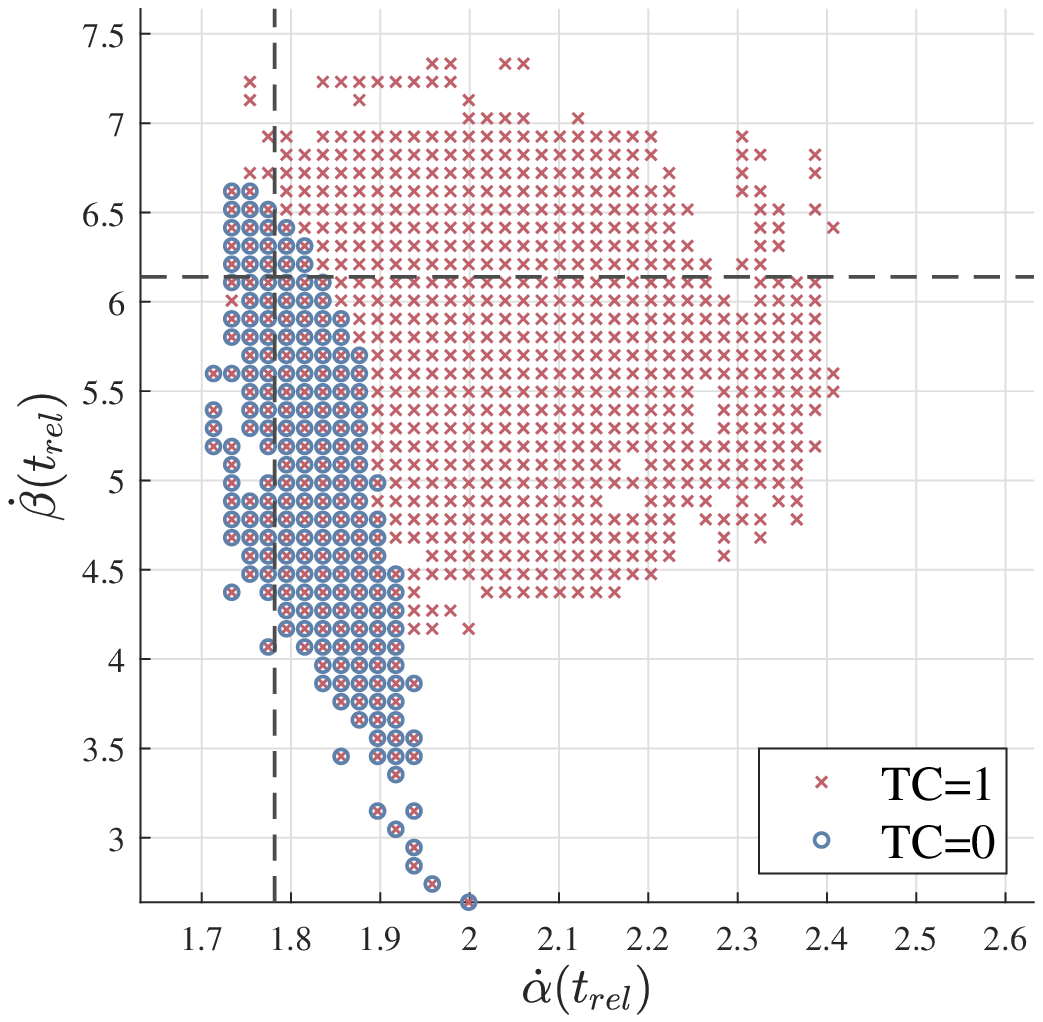}
    \end{subfigure}
    \caption{This figure shows how the initial releasing point influences Mon$\chi$'s ability to land. Successful attempts \textit{with} trajectory correction are marked in red, and successful attempts \textit{without} trajectory correction are marked in blue.}
    \label{fig:region_plot}
\end{figure}

\subsection{Flying Trapeze Act Simulation}
The entire system is tested in the simulation. Various parameters have been altered in this experiment to reflect uncertainty in the real world. Those parameters are the robot's initial position, inertial properties, and damping coefficients. First, inertial properties, e.g., mass, the moment of inertia, and the center-of-mass position, are correlated. For that reason, the function that can calculate uncertainties according to the fabrication tolerance of each part is written. Results are shown in \cref{tab:exp_param}. Unfortunately, the damping coefficient cannot be obtained yet since the system has not been realized; hence, relatively high uncertainty is selected.  

\begin{table}[htpb]
    \begin{center}
        \caption{\label{tab:exp_param}Robot Parameters for the Experiment}
        \renewcommand{\arraystretch}{1.5}
        \begin{tabular}{ | m{1.6cm} | m{4.0cm} | } 
            \hline
            Parameter & Value  \\ 
            \hline
            $m_1$ & $\SI{1.5790 (76)}{\kilo\gram}$ \\ 
            \hline
            $m_2$ & $\SI{1.4370 (400)}{\kilo\gram}$ \\ 
            \hline
            $r_1$ & $\SI{0.1443 (52)}{\meter}$ \\
            \hline
            $r_2$ & $\SI{0.1269 (33)}{\meter}$ \\
            \hline
            $\kappa_1$ & $\SI{-0.0369 (1)}{\radian}$ \\
            \hline
            $\kappa_2$ & $\SI{0.0001 (23)}{\radian}$ \\
            \hline
            $I_{1}$ & $\SI{0.0375 (13)}{\kilo\gram\square\meter}$ \\ 
            \hline
            $I_{2}$ & $\SI{0.0237 (9)}{\kilo\gram\square\meter}$ \\ 
            \hline
            $\sigma_\alpha$ & $\SI{0.050 (10)}{\newton\meter\per\radian\per\second}$ \\ 
            \hline
            $\sigma_\beta$ & $\SI{0.200 (40)}{\newton\meter\per\radian\per\second}$ \\ 
            \hline
        \end{tabular}
    \end{center}
\end{table}

\begin{figure}[htbp]
    \centering
    \includegraphics[height=14em]{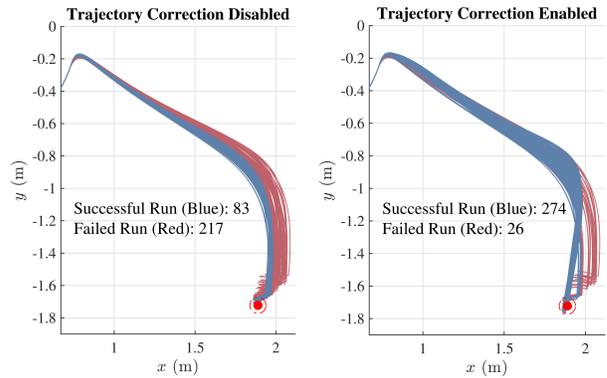}
    \caption{The gripper trajectory of a controller \textit{without} trajectory correction (left) and \textit{with} trajectory correction (right).}
    \label{fig:sim_results}
\end{figure}

In the simulation, contacts between objects were also added to simulate effects from friction and contact forces between objects, especially between a robot's gripper and trapeze's handle, which affects the center-of-mass trajectory ($\mathbf{p}_0^c$). Three hundred test scenarios are generated for each simulation setting based on parameters in \cref{tab:exp_param}. Other than parameters in \cref{tab:exp_param}, uncertain initial position is also added, with a maximum magnitude of $\pm$\SI{0.01}{\radian} for both $\alpha_0$ and $\beta_0$. The result is shown in \cref{tab:full_sim_result}. Comparison of the gripper trajectory between a controller with and without trajectory correction are plotted in Fig. \ref{fig:sim_results}. Note that the successful trajectories on the right of the Fig. \ref{fig:sim_results} (blue) can be grouped into two categories. First are the trajectories that should have missed the target but were corrected by the trajectory correction module. And there are a few that did not trigger the trajectory correction module since the releasing states ($\mathbf{p}_0^c(t_{rel})$ and $\mathbf{v}_0^c(t_{rel})$) did not deviate much from the baseline. The trajectories in the latter group are similar to some of the successful trajectories on the left plot.

\begin{table}[htpb]
    \begin{center}
        \caption{\label{tab:full_sim_result}Simulation Result}
        \renewcommand{\arraystretch}{1.5}
        \begin{tabular}{ | m{3.6cm} | c | c | c |} 
            \hline
            Controller Configuration & Success & Fail & Success\%\\ 
            \hline
            Open Loop ($u(t) = u^*(t)$) & 1 & 299 & $\SI{0.33}{\percent}$\\ 
            \hline
            Posture Control and TC = 0 & 83 & 217 & $\SI{27.67}{\percent}$\\ 
            \hline
            Posture Control and TC = 1 & 274 & 26 & $\SI{91.33}{\percent}$\\ 
            \hline
        \end{tabular}
    \end{center}
\end{table}

%% file: root.bbl
\begin{thebibliography}{1}
\providecommand{\url}[1]{#1}
\csname url@samestyle\endcsname
\providecommand{\newblock}{\relax}
\providecommand{\bibinfo}[2]{#2}
\providecommand{\BIBentrySTDinterwordspacing}{\spaceskip=0pt\relax}
\providecommand{\BIBentryALTinterwordstretchfactor}{4}
\providecommand{\BIBentryALTinterwordspacing}{\spaceskip=\fontdimen2\font plus
\BIBentryALTinterwordstretchfactor\fontdimen3\font minus
  \fontdimen4\font\relax}
\providecommand{\BIBforeignlanguage}[2]{{%
\expandafter\ifx\csname l@#1\endcsname\relax
\typeout{** WARNING: IEEEtran.bst: No hyphenation pattern has been}%
\typeout{** loaded for the language `#1'. Using the pattern for}%
\typeout{** the default language instead.}%
\else
\language=\csname l@#1\endcsname
\fi
#2}}
\providecommand{\BIBdecl}{\relax}
\BIBdecl

\bibitem{pope-2018}
\BIBentryALTinterwordspacing
M.~T. Pope, S.~Christensen, D.~Christensen, A.~Simeonov, G.~Imahara, and
  G.~Niemeyer, ``Stickman: Towards a human scale acrobatic robot,'' in
  \emph{2018 {IEEE} International Conference on Robotics and Automation
  ({ICRA})}.\hskip 1em plus 0.5em minus 0.4em\relax {IEEE}, May 2018, pp.
  2134--2140. [Online]. Available:
  \url{https://doi.org/10.1109/icra.2018.8462836}
\BIBentrySTDinterwordspacing

\bibitem{nakanishi-2000}
\BIBentryALTinterwordspacing
J.~Nakanishi and T.~Fukuda, ``A leaping maneuvre for a brachiating robot,'' in
  \emph{Proceedings 2000 {ICRA}. Millennium Conference. {IEEE} International
  Conference on Robotics and Automation. Symposia Proceedings (Cat.
  No.00CH37065)}.\hskip 1em plus 0.5em minus 0.4em\relax {IEEE}, 2000, pp.
  2822--2827. [Online]. Available:
  \url{https://doi.org/10.1109/robot.2000.846455}
\BIBentrySTDinterwordspacing

\bibitem{wan-2015}
\BIBentryALTinterwordspacing
D.~Wan, H.~Cheng, G.~Ji, and S.~Wang, ``Non-horizontal ricochetal brachiation
  motion planning and control for two-link bio-primate robot,'' in \emph{2015
  {IEEE} International Conference on Robotics and Biomimetics ({ROBIO})}.\hskip
  1em plus 0.5em minus 0.4em\relax {IEEE}, Dec. 2015, pp. 19--24. [Online].
  Available: \url{https://doi.org/10.1109/robio.2015.7407033}
\BIBentrySTDinterwordspacing

\bibitem{dubowsky-1993}
\BIBentryALTinterwordspacing
S.~Dubowsky and E.~Papadopoulos, ``The kinematics, dynamics, and control of
  free-flying and free-floating space robotic systems,'' \emph{{IEEE}
  Transactions on Robotics and Automation}, vol.~9, no.~5, pp. 531--543, 1993.
  [Online]. Available: \url{https://doi.org/10.1109/70.258046}
\BIBentrySTDinterwordspacing

\bibitem{hargraves-1987}
\BIBentryALTinterwordspacing
C.~Hargraves and S.~Paris, ``Direct trajectory optimization using nonlinear
  programming and collocation,'' \emph{Journal of Guidance, Control, and
  Dynamics}, vol.~10, no.~4, pp. 338--342, Jul. 1987. [Online]. Available:
  \url{https://doi.org/10.2514/3.20223}
\BIBentrySTDinterwordspacing

\bibitem{barry-2017}
\BIBentryALTinterwordspacing
A.~J. Barry, P.~R. Florence, and R.~Tedrake, ``High-speed autonomous obstacle
  avoidance with pushbroom stereo,'' \emph{Journal of Field Robotics}, vol.~35,
  no.~1, pp. 52--68, Sep. 2017. [Online]. Available:
  \url{https://doi.org/10.1002/rob.21741}
\BIBentrySTDinterwordspacing

\bibitem{moore-2016}
\BIBentryALTinterwordspacing
J.~Moore, K.~C. Wolfe, M.~S. Johannes, K.~D. Katyal, M.~P. Para, R.~J. Murphy,
  J.~Hatch, C.~J. Taylor, R.~J. Bamberger, and E.~Tunstel, ``Nested marsupial
  robotic system for search and sampling in increasingly constrained
  environments,'' in \emph{2016 {IEEE} International Conference on Systems,
  Man, and Cybernetics ({SMC})}.\hskip 1em plus 0.5em minus 0.4em\relax {IEEE},
  Oct. 2016. [Online]. Available:
  \url{https://doi.org/10.1109/smc.2016.7844578}
\BIBentrySTDinterwordspacing

\bibitem{farzan-2019}
\BIBentryALTinterwordspacing
S.~Farzan, A.-P. Hu, E.~Davies, and J.~Rogers, ``Feedback motion planning and
  control of brachiating robots traversing flexible cables,'' in \emph{2019
  American Control Conference ({ACC})}.\hskip 1em plus 0.5em minus 0.4em\relax
  {IEEE}, Jul. 2019. [Online]. Available:
  \url{https://doi.org/10.23919/acc.2019.8814894}
\BIBentrySTDinterwordspacing

\bibitem{xin-2004}
\BIBentryALTinterwordspacing
X.~Xin, T.~Mita, and M.~Kaneda, ``The posture control of a two-link free flying
  acrobot with initial angular momentum,'' \emph{{IEEE} Transactions on
  Automatic Control}, vol.~49, no.~7, pp. 1201--1206, Jul. 2004. [Online].
  Available: \url{https://doi.org/10.1109/tac.2004.831093}
\BIBentrySTDinterwordspacing

\end{thebibliography}
